\documentclass[11pt,a4paper]{article}
\usepackage[hyperref]{acl2019}
\usepackage{times}
\usepackage{latexsym}
\usepackage{url}
\usepackage{tikz}
\usepackage{wrapfig}
\usepackage{pgfplots}
\usepackage{amsmath,amsfonts,amssymb}
\usepackage{graphicx}
\usepackage{caption}
\usepackage{subcaption}

\aclfinalcopy % Uncomment this line for the final submission
\title{An Unsupervised Approach for Aspect Category Detection Using Soft Cosine Similarity Measure}

\author{Erfan Ghadery\thanks{\hspace{0.2cm} Equal Contribution.}, Sajad Movahedi\footnotemark[1], Heshaam Faili, Azadeh Shakery\\
University of Tehran\\
Tehran, Iran\\
\{erfan.ghadery, s.movahedi, hfaili, shakery\}@ut.ac.ir\\
}

\date{}

\begin{document}
\maketitle
\setcounter{figure}{1} 
\begin{abstract}
Aspect category detection is one of the important and challenging subtasks of aspect-based sentiment analysis. Given a set of pre-defined categories, this task aims to detect categories which are indicated implicitly or explicitly in a given review sentence. Supervised machine learning approaches perform well to accomplish this subtask. Note that, the performance of these methods depends on the availability of labeled train data, which is often difficult and costly to obtain. Besides, most of these supervised methods require feature engineering to perform well. In this paper, we propose an unsupervised method to address aspect category detection task without the need for any feature engineering. Our method utilizes clusters of unlabeled reviews and soft cosine similarity measure to accomplish aspect category detection task. \footnote{We have made our code available at https://github.com/erfan-ghadery/Unsupervised-Aspect-Category-Detection.} Experimental results on SemEval-2014 restaurant dataset shows that proposed unsupervised approach outperforms several baselines by a substantial margin.
\end{abstract}

\section{Introduction}
User-generated reviews are valuable resources for both companies and potential customers. These reviews can help online retail companies in recognizing their weaknesses and strengths, and facilitate the decision process for consumers. However, it is impractical and tedious to read such a huge amount of reviews one by one manually. Therefore, the need for an automatic system that processes a huge amount of reviews and provides a summary of these different reviews in a suitable form seems necessary.             
\par Given a list of pre-defined aspect categories (e.g. `food' and `price' in restaurant domain), aspect category detection aims to assign a subset of these categories to review sentences. SemEval 2014 Task 4 \cite{semeval2014} tried to tackle this task by providing datasets for multiple domains, including restaurant and laptop reviews.
\par Previous works in the literature mostly are based on supervised machine learning approaches \cite{ccetin2016tgb}\cite{kiritchenko2014nrc}\cite{2014unitor}. Although these methods perform well, they need annotated train data which are time-consuming and expensive to obtain. Thus, unsupervised approaches maybe be a good choice, especially for low resource languages.    
\par Soft cosine measure is a similarity measure that assesses the similarity between two sentences, even when they have no words in common\cite{sidorov2014soft}. For aspect category detection, our method utilizes soft cosine similarity. Firstly, we cluster a set of unlabeled review sentences into k cluster. Clustering is performed based on the Euclidean distance between the average of their word embeddings. Our motivation for using the cluster of sentences is based on the intuition that sentences in the same cluster share similar information about categories they belong to. The similarity between a given sentence and a pre-defined category is defined as the soft cosine similarity between sentence and a set of manually selected seed words corresponding to that category. So, the similarity values can give us information about categories that sentence belongs to. We also define similarity between a cluster and a category as averaging the similarity scores of the sentences in the cluster. Finally, given a test review sentence, scores obtained for the sentence and the nearest cluster to it are interpolated. These final scores are normalized and used to detect the categories mentioned in the sentence. If the similarity of a category surpasses a threshold, it is assigned to the sentence.
\par We evaluate our method on SemEval-2014 restaurant dataset. Experimental results show that the proposed method outperforms several baselines by a high margin.

\section{Related works}
Early works for addressing aspect extraction relied on approaches such as identifying frequent nouns and noun phrases using association rule mining, dependency relations, and lexical patterns\cite{hu2004mining}\cite{qiu2011opinion}\cite{popescu2007}. The SemEval workshops, over the course of three years, has included aspect-based sentiment analysis in their competitions. One of the subtasks introduced during SemEval is aspect category detection, which our proposed method is going to address. Most of the supervised approaches proposed to address this subtask utilizing machine learning algorithms and train a set of the one-vs-all classifier using hand-crafted features \cite{kiritchenko2014nrc}, \cite{xenos2016aueb}, \cite{toh2016nlangp}.

\par There are only a few unsupervised approaches to address the aspect category detection subtask in the literature. In \cite{he2017Attention} authors trained a network similar to auto-encoder with attention mechanism to attend to aspect-related words. \cite{v32014} propose to use a double propagation technique to mine rules based on dependency relations for finding aspect terms of each category. Also, irrelevant aspect terms were pruned using stop words and the PageRank algorithm on a graph-based approach. \cite{schouten2018supervised} proposed an unsupervised method called `spreading activation' that performs association rule mining, using a set of seed words and a co-occurrence matrix between words to form a co-occurrence digraph to detect aspect categories. One of the shortcomings of this method is the need for tuning multiple parameters. Unlike this method, our proposed method only requires a small number of parameter (a threshold, the number of clusters, and the interpolation coefficient).

\section{Proposed Method}

This section describes our proposed unsupervised method. In the following subsections, we will discuss each of the main components in detail.
\subsection{Manual Selection of Category Seed Words}
We manually select a set of 5 seed words for each category (20 in total) to represent the category. Because the anecdotes/miscellaneous category is very unspecific and abstract, we didn't choose any seed words for it\cite{schouten2018supervised}. For this aspect, following \cite{schouten2018supervised}, we would assign this category only to sentences that were not assigned any other category. The set of seed words for each category can be seen in Table~\ref{tab1}.

\begin{table}[h!]
\small
\caption{List of the seed words for each category.}\label{tab1}
    \centering
    \scalebox{0.85}{
    \begin{tabular}{c|l}
    \centering
    Category & Seed Words \\ \hline
    food & food, delicious, menu, fresh, tasty \\ \hline
    service & service, staff, friendly, attentive, manager \\ \hline
    price & price, cheap, expensive, money, affordable \\ \hline
    ambience & ambience, atmosphere, decor, romantic, loud \\
\end{tabular}}
\end{table}

\subsection{Sentence similarity}
In order to find the similarity score of a given sentence compares to a category, we utilize soft cosine measure. For each category, we define the similarity of the given sentence to that category as the average of soft cosine similarity values between the sentence and each of the seed words belonging to that category. Let $x$ be a given sentence and $a_i$ be the $i$-th category. We define the $sentSim_{a_i}(x)$ to be the similarity value between $a_i$ and $x$ as seen in equation~\ref{eq1}.
\begin{equation} 
\label{eq1}
{sentSim_{a_{i}}(x)} = \frac{\sum_{\substack{i=1}}^{|s|} softcossim(x,s_{i})}{|s|}
\end{equation}

\par As stated in \cite{zamani}, the sigmoid function can have a discriminating effect on the similarity values obtained from similarity measures like cosine similarity. To make the similarity values more discriminating, we transferred the similarity values obtained in the previous step via the sigmoid function as seen in equation~\ref{eq2}. 
\begin{equation} \label{eq2}
{sentScore_{a_{i}}(x)} = \frac{e^{sentSim_{a_{i}}(x)}}{1 + e^{sentSim_{a_{i}}(x)}}
\end{equation}
\par Now, for each sentence we have a vector $sentScore \in \mathbb{R}^c$, which c is the number of categories and each element represents the similarity score between the sentence and a pre-defined category.

\subsection{Cluster similarity}
A set of unlabeled sentences are acquired from the Yelp dataset challenge \footnote{https://www.yelp.com/dataset/challenge}. In order to decrease noise samples and since the precision is a more important factor than recall in acquiring true sentences, only sentences that contain at least one of the category names (eg. 'food', 'service') are selected. Using k-means clustering algorithm, these unlabeled sentences are clustered into k clusters. Clustering is done based on the Euclidean distance between the average of word embedding of sentence words, where word embeddings are trained on the Yelp dataset using Continuous Bag of Words (CBOW) algorithm. For each cluster, a similarity value per category is calculated as shown in equation~\ref{eq3}:

\begin{equation} \label{eq3}
{clustSim_{a_{i}}(c_{k})} = \frac{\sum_{x \in c_{k}} sentSim_{a_{i}}(x)}{|c_{k}|} 
\end{equation}

\noindent where $c_{k}$ is the $k$-th cluster. Similar to sentence similarity, we also utilize sigmoid function in here for it's discriminating effect.

\begin{equation} \label{eq4}
{clustScore_{a_{i}}(x)} = \frac{e^{clustSim_{a_{i}}(x)}}{1 + e^{clustSim_{a_{i}}(x)}}
\end{equation}

\par Therefore, for each cluster we have a vector \begin{math}clustScore \in \mathbb{R}^c\end{math}, which c is the number of categories and each element represents the similarity between cluster samples and one of the pre-defined categories.

\subsection{Assign aspect categories}
\par Given a test review sentence, first, we calculate the sentScore vector for the sentence, then, this vector is interpolated with the clustScore corresponding to the nearest cluster to the given sentence, as shown in equation~\ref{eq5}. The nearest cluster is found by finding the closest centroid to the sentence based on the Euclidean distance between the average of the word embeddings of the sentence and the cluster centroid.

\begin{equation}
    \small
    \label{eq5}
    {score}= \alpha sentScore_{N} + (1-\alpha) clustScore_{N}
\end{equation}

\noindent where $sentScore_{N}$ and $clustScore_{N}$ are L2-Normalized vectors of sentence and its cluster respectively. Categories that exceed a threshold are assigned to the sentence. We find the optimal threshold by a simple linear search.

\section{Experiments}
We train word embeddings on the Yelp challenge unlabeled dataset. Word embedding size is set to 300 and CBOW model \cite{mikolov2013efficient} with default parameters is used for training the embedding vectors. Removing stop-words and tokenizing sentences are performed as the pre-processing step using NLTK package \cite{bird2009natural}. For implementing soft cosine similarity measure and word embedding training, we used gensim package \cite{rehurek_lrec}. A simple linear search is performed to find the optimal hyperparameters \begin{math}\alpha\end{math} and \begin{math}K\end{math} - the interpolation coefficient between sentence score and its cluster score, and the number of clusters for k-means clustering respectively. We set the parameter \begin{math}\alpha\end{math} to 0.7 and the number of clusters to 17 to obtain the best result in our method.

\begin{figure}[t]
    \addtocounter{figure}{-1}
    \centering
    \begin{subfigure}[t]{0.4\textwidth}
    \centering
        \begin{tikzpicture}
    \begin{axis}[
        width=1.0\textwidth,
        height=1.0\textwidth,
        xlabel={$\alpha$},
        ylabel={\% F1-score},
        xmin=-0.1, xmax=1.1,
        ymin=60, ymax=80,
        xtick={0.0, 0.2, 0.4, 0.6, 0.8, 1.0},
        ytick={60,65,70, 75, 80},
        legend pos=north west,
        ymajorgrids=false,
        grid style=false,
    ]
    \addplot[
        color=black,
        mark=square,
        ]
        coordinates {
        (0.0, 62.136986301369866)
        (0.1, 65.76234400434075)
        (0.2, 68.71297792137857)
        (0.3, 71.48981779206859)
        (0.4, 74.09766454352442)
        (0.5, 75.27675276752768)
        (0.6, 76.34237107921318)
        (0.7, 76.67020148462353)
        (0.8, 76.42782969885774)
        (0.9, 75.55099948744234)
        (1.0, 71.79487179487181)
        };
    \end{axis}
    \end{tikzpicture}
    \subcaption{Interpolation coefficient $\alpha$ sensitivity}
    \end{subfigure}
    \begin{subfigure}[t]{0.4\textwidth}
    \centering
    \begin{tikzpicture}
    \begin{axis}[
    width=1.0\textwidth,
        height=1.0\textwidth,
        xlabel={$k$},
        ylabel={\% F1-score},
        xmin=0, xmax=32,
        ymin=60, ymax=80,
        xtick={0, 4, 8, 12, 16, 20, 24, 28, 32},
        ytick={60,65,70, 75, 80},
        legend pos=north west,
        ymajorgrids=false,
        grid style=false,
    ]
 
    \addplot[
        color=black,
        mark=.,
        ]
        coordinates {
        (1, 72.85936689154124)
        (2, 73.06694343539182)
        (3, 73.09697601668404)
        (4, 72.3157894736842)
        (5, 75.16918271733472)
        (6, 75.5741127348643)
        (7, 76.15062761506276)
        (8, 76.28004179728319)
        (9, 76.42782969885774)
        (10, 75.46777546777548)
        (11, 75.31380753138075)
        (12, 75.53693033001572)
        (13, 74.59740259740259)
        (14, 75.93078133193497)
        (15, 75.8909853249476)
        (16, 76.97127937336813)
        (17, 76.98744769874477)
        (18, 76.97095435684648)
        (19, 75.88726513569937)
        (20, 76.02704108164326)
        (21, 76.73935617860852)
        (22, 75.72916666666667)
        (23, 75.7686294945284)
        (24, 75.33960292580981)
        (25, 75.27216174183515)
        (26, 75.83333333333333)
        (27, 75.22077922077922)
        (28, 73.97827211588205)
        (29, 73.46094154164511)
        (30, 74.45708376421922)
        };
    \end{axis}
    \end{tikzpicture}
    \subcaption{The effect of number of clusters k.}
    \end{subfigure}
    \caption{The effect of hyperparameters.}
\end{figure}
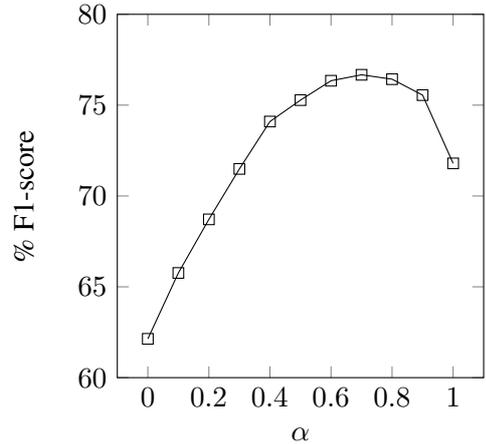
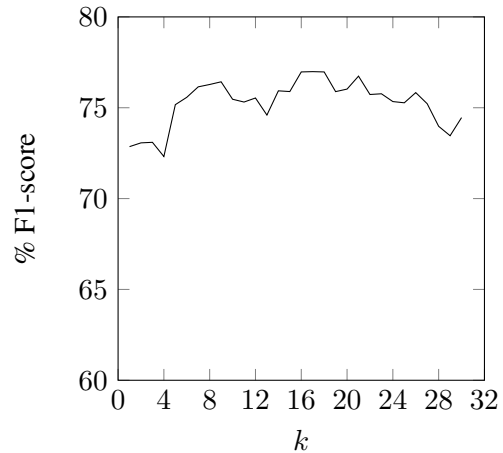

\subsection{Results and Analysis}
Micro F1-score, precision, and recall of all the category labels are used as evaluation metrics and performance of the proposed method evaluated on test data from SemEval-2014\cite{semeval2014}. The test data contains 800 test review sentence in the restaurant domain.
\par The proposed method is compared to the following baselines. \textbf{Random} baseline assigns a random category based on the frequency of the categories appearing in the train data. \textbf{Majority} baseline assigns the two most common categories in the train data ('food' and 'anecdotes/miscellaneous') to each of the test sentences. \textbf{COMMIT-P1WP3} \cite{schouten2014commit} baseline is a co-occurrence based method. After constructing a co-occurrence matrix between the words and the pre-defined categories, this method calculates the probability of a given sentence belonging to each category based on this matrix. \textbf{V3} \cite{v32014} baseline utilizes WordNet similarity to compare the detected aspect terms to representative terms of each category. \textbf{SemEval-2014 baseline} \cite{semeval2014} is the baseline provided by the SemEval 2014 which uses a simple k-nearest neighbor classifier to detect aspect categories. In the \textbf{Spreading Activation} baseline \cite{schouten2018supervised}, using a set of seed words and a co-occurrence matrix similar to \cite{schouten2014commit}, a set of association rules are mined, associating categories to terms appearing in the data. The aspect category detection task is then performed using these association rules.
\par Table~\ref{tab2} shows that among baseline methods, Spread Activation, an unsupervised approach, performs the best with F1-score 67\%. Our unsupervised method outperforms Spread Activation by 9.98\% in terms of F1-score and similarly improves over SemEval-2014 baseline, V3, COMMIT-P1WP3, Majority, and Random baselines by 13.08\%, 16.78\%, 17.68\%, 27.31\% and 47.30\% respectively. These results clearly demonstrate the effectiveness of the proposed unsupervised method in aspect category detection task.

{\small
\begin{table}[h]
\caption{The result of baseline methods compared to our method. The COMMIT-P1EP3 method and the SemEval 2014 method are supervised methods (S) and the V3 and Spreading Activation method are unsupervised methods (U).}\label{tab2}
    \centering
    \scalebox{0.8}{
    \begin{tabular}{l|c|c|c}
    \centering
    Method & precision & recall & $F_{1}$ \\ \hline
    Random  & 34\% & 26.34\% & 29.68\%  \\
    Majority  & 40.75\% & 63.60\% & 49.67\%  \\
    COMMIT-P1WP3 $^{S}$  & 63.3\% & 55.8\% & 59.3\%  \\
    V3 $^{U}$ & 63.3\% & 56.9\% & 60.2\%  \\
    SemEval-2014 baseline $^{S}$ & - & - & 63.9\%  \\
    Spreading Activation $^{U}$ & 69.5\% & 64.7\% & 67.0\%  \\
    \hline
    \textbf{Our method} & \textbf{82.97\%} & \textbf{71.80\%} & \textbf{76.98\%}  \\
\end{tabular}}
\end{table}
}

\subsection{Hyperparameter tuning}
Figure 1 (a) plots the sensitivity of our system to the interpolation coefficient $\alpha$ in equation 5. According to the curve, to achieve the best performance a higher weight should be given to sentence scores itself, which indicates that it plays the main role in finding corresponding categories. The results of the experiments prove our intuition that interpolating the cluster scores and the sentence scores improves the classification performance. The reason behind this improvement can be the sentences that contain a seed word or a semantically close word to the seed words of a category. Such sentences will improve the scores of other sentences in the cluster for that category. The best result was obtained at $\alpha = 0.7$. We also conducted experiments on the number of clusters in the k-means clustering algorithm. Figure 1 (b) provides the results of these experiments. We found the optimum value for k to be 17. As can be seen in the figure, the result of our method tends to deteriorate with k larger than 17. This can be due to the lowering consistency of clusters with larger k. Also, for the k smaller than 9, the same thing can happen.

\section*{Conclusions}
In this paper, we propose an unsupervised approach for aspect category detection that utilizes soft cosine similarity and the well-known k-means clustering to detect categories belonging to a review sentence. Experimental results on SemEval-2014 benchmark dataset show the effectiveness of our method compared to several supervised and unsupervised baseline methods.

\bibliography{acl2019}
\bibliographystyle{acl_natbib}

\end{document}